\documentclass[10pt,twocolumn,letterpaper]{article}

\usepackage{iccv}
\usepackage{times}
\usepackage{epsfig}
\usepackage{graphicx}
\usepackage{amsmath}
\usepackage{amssymb}

\usepackage[pagebackref=true,breaklinks=true,letterpaper=true,colorlinks,bookmarks=false]{hyperref}

\iccvfinalcopy %

\def\iccvPaperID{} %

\usepackage{nicefrac}       %
\usepackage{microtype}      %

\usepackage{blindtext}
\usepackage{float}
\usepackage{enumitem}
\usepackage[table]{xcolor}
\usepackage{tabulary,multirow,overpic}
\usepackage{verbatim}
\usepackage{color}
\usepackage{dblfloatfix}
\usepackage{pifont}%
\usepackage[accsupp]{axessibility}  %
\usepackage{makecell}

\usepackage{marvosym}

\usepackage{dblfloatfix}
\usepackage{nicefrac}

\usepackage{relsize}

\usepackage{listings}
\definecolor{linkcolor}{RGB}{255,0,0}
\definecolor{urlcolor}{RGB}{255,105,180}
\definecolor{citecolor}{RGB}{66,168,235}
\definecolor{codegreen}{rgb}{0,0.5,0}
\definecolor{codeblue}{rgb}{0.25,0.5,0.5}
\definecolor{codegray}{rgb}{0.6,0.6,0.6}

\newlength\savewidth\newcommand\shline{\noalign{\global\savewidth\arrayrulewidth
		\global\arrayrulewidth 1pt}\hline\noalign{\global\arrayrulewidth\savewidth}}

\newcolumntype{x}[1]{>{\centering\arraybackslash}p{#1pt}}
\newcolumntype{y}[1]{>{\raggedright\arraybackslash}p{#1pt}}
\newcolumntype{z}[1]{>{\raggedleft\arraybackslash}p{#1pt}}

\definecolor{baselinecolor}{gray}{.92}

\definecolor{demphcolor}{gray}{.2}

\definecolor{demphcolorinline}{gray}{.3}

\definecolor{demphcolor1}{gray}{.6}

\newcommand{\gr}[1]{{\textcolor{gray}{#1}}}

\renewcommand{\paragraph}[1]{\vspace{1.25mm}\noindent\textbf{#1}}

\newcommand{\app}{\raise.17ex\hbox{$\scriptstyle\sim$}}

\newcommand{\authorskip}{\hspace{5mm}}

\usepackage{marvosym}

\usepackage[capitalize]{cleveref}
\crefname{section}{Sec.}{Secs.}
\Crefname{section}{Section}{Sections}
\Crefname{table}{Table}{Tables}
\crefname{table}{Table}{Tables}

\newcommand{\myparagraph}[1]{{\vspace{.1em} \noindent \bf #1}}
\def\Ours{{SegGPT}\xspace}

\usepackage{caption}
\usepackage{subcaption}

\begin{document}

\title{\Ours: Segmenting Everything In Context}

\author{Xinlong Wang\textsuperscript{1}\thanks{Equal contribution. Correspondence to \textit{xinlong.wang96@gmail.com}. 
} 
\authorskip Xiaosong Zhang\textsuperscript{1}$^*$ \authorskip Yue Cao\textsuperscript{1}$^*$ \authorskip Wen Wang\textsuperscript{2} \authorskip  
Chunhua Shen\textsuperscript{2} \authorskip Tiejun Huang\textsuperscript{1,3} \\[2mm]
{
\fontsize{10.4pt}{9.84pt}\selectfont
\textsuperscript{1} Beijing Academy of Artificial Intelligence \hspace{5.7mm} 
\textsuperscript{2} Zhejiang University \hspace{5.7mm} \textsuperscript{3} Peking University}\\[1.5mm]
{
\fontsize{9.4pt}{9.84pt}\selectfont 
Code \& Demo: \url{https://github.com/baaivision/Painter}
}
}

\input figures/teaser.tex

\maketitle
\ificcvfinal\thispagestyle{empty}\fi

\begin{abstract}
   We present \Ours, a generalist model for segmenting everything in context. We unify 
   various 
   segmentation tasks into a generalist in-context learning framework that accommodates different kinds of segmentation data by transforming them into the same format of images. The training of \Ours is formulated as an in-context coloring problem with random color mapping for each data sample. The objective is to accomplish diverse tasks according to the context, rather than relying on specific colors. After training, \Ours can perform arbitrary segmentation tasks in images or videos via in-context inference, such as object instance, stuff, part, contour, and text. \Ours is evaluated on a broad range of tasks, including few-shot semantic segmentation, video object segmentation, semantic segmentation, and panoptic segmentation. Our results show strong capabilities in segmenting in-domain and out-of-domain targets, either qualitatively or quantitatively. 
\end{abstract}

\section{Introduction}

Segmentation is one of the most fundamental problems in computer vision, which aims to localize and re-organize meaningful concepts at the pixel level, \eg, foreground, category, object instance, \etc.
During recent years, we have witnessed great progress in developing more accurate and faster algorithms for various segmentation tasks, such as foreground segmentation~\cite{stauffer1999adaptive}, interactive segmentation~\cite{xu2016deepinteractive, mahadevanitis}, semantic segmentation~\cite{long2015fully,lin2017refinenet,pspnet,ranftl2021vision}, instance segmentation~\cite{he2017mask,de2017semantic,yolact,wang2021SOLO}, and panoptic segmentation~\cite{panopticfpn,carion2020detr,mask2former}.

However, these specialist segmentation models are limited to specific tasks, classes, granularities, data types, \etc. 
A new model has to be trained when adapting to a different setting, \eg, to segment a novel concept, or to segment objects in videos instead of images. This requires expensive annotation efforts and is not sustainable for a large number of segmentation tasks.

In this work, we aim to train a single model that is capable of solving diverse and unlimited segmentation tasks. The main challenges are twofold:  (1) to incorporate those very different data types in training, \eg, part, semantic, instance, panoptic, person, medical image, aerial image, \etc; (2) to design a generalizable training scheme that differs from conventional multi-task learning, which is flexible on task definition and is capable of handling out-of-domain tasks.

To address these challenges, we present \Ours, a generalist model for segmenting everything in context. We view segmentation as a general format for visual perception and unify different segmentation tasks into a generalist in-context learning framework~\cite{painter}. This framework accommodates different kinds of segmentation data by transforming them into the same format of images. The training of \Ours is formulated as an in-context coloring problem with random color mapping for each data sample. The objective is to color the corresponding areas, such as classes, object instances, parts, \etc, only according to the context. By using a random coloring scheme, the model is forced to reference contextual information to complete the assigned task, instead of relying on specific colors. This allows for a more flexible and generalizable approach to training. The remaining parts of training keep the same as \cite{painter} using a vanilla ViT~\cite{vaswani2017attention} and a simple smooth-$\ell_1$~\cite{girshick2015fast} loss.

After training, \Ours is able to perform diverse segmentation tasks in images or videos given a few examples via in-context inference, such as object instance, stuff, part, contour, text, \etc. To effectively ensemble multiple examples in context, we propose a simple yet effective context ensemble strategy, the feature ensemble, which can help the model benefit from the multi-example prompting setting.
Additionally, \Ours can conveniently serve as a specialist model without updating the model parameters, by tuning a specific prompt for a specialized use case, such as in-domain ADE20K semantic segmentation.

Our \textbf{main contributions} are as follows. 
(1) For the first time, we demonstrate a single generalist model capable of performing
a  diverse set of  segmentation tasks automatically.
(2) We evaluate the pre-trained \Ours on a broad range of tasks directly, \ie, without fine-tuning, including few-shot semantic segmentation, video object segmentation, semantic segmentation, and panoptic segmentation. 
(3) 
Our results show strong capabilities in segmenting in-domain and out-of-domain targets, either qualitatively or quantitatively. 

However, this work does not aim to claim new state-of-the-art results or outperform existing specialist methods across all benchmarks, 
as we believe that this may not be the responsibility of a general-purpose model.

\begin{figure*}
    \centering
    \includegraphics[width=0.95\linewidth]{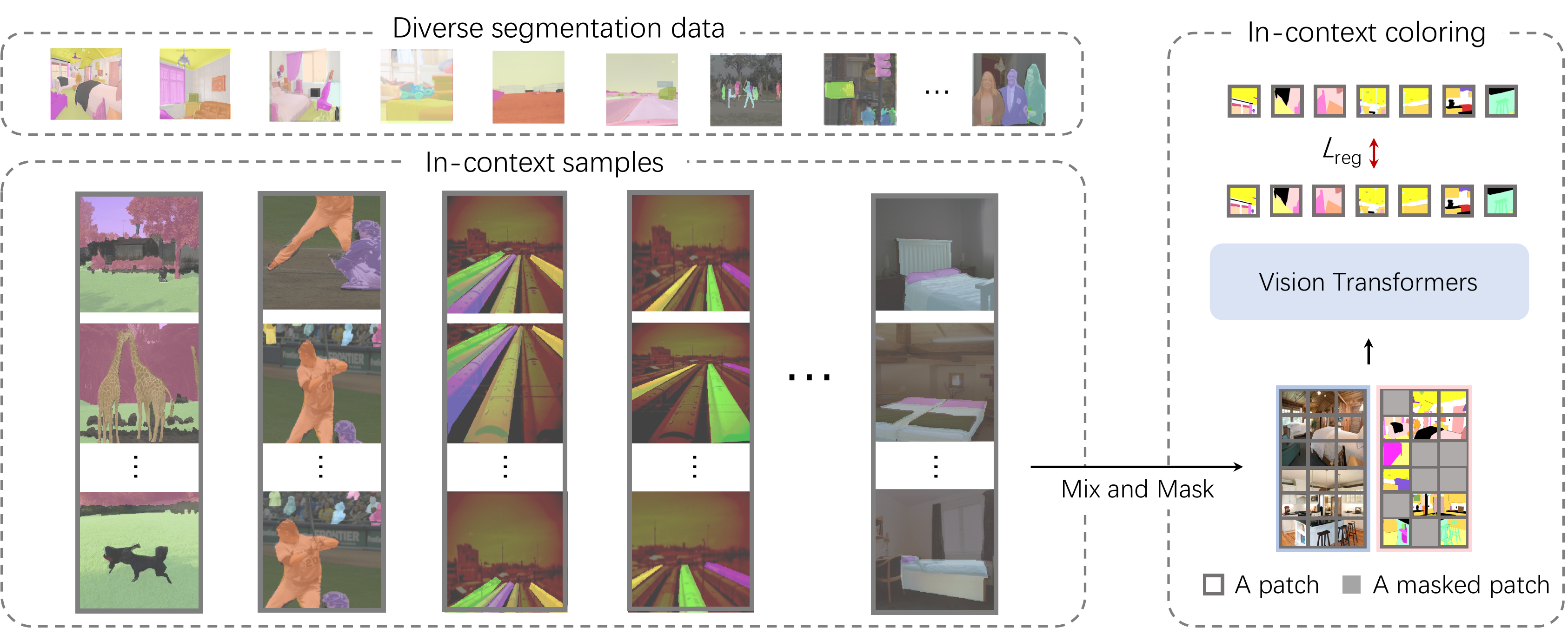}
    \caption{Illustration of overall training framework of \Ours. We incorporate diverse segmentation data, including part, semantic, instance, panoptic, person, medical image, and aerial image segmentation, and transform them into the same format of images. We generate in-context samples that share similar contexts on-the-fly, \eg, the overlapped colors shown in each column, which indicate the same category or the same instance. We adopt a general Painter~\cite{painter} framework with in-context coloring as the training objective and a random coloring scheme for more flexible and generalizable training.}
    \label{fig:method_coloring}
\end{figure*}

\section{Related Work}
\subsection{Visual Segmentation}

Segmentation is a fundamental problem in computer vision that involves localizing and organizing meaningful concepts at the pixel level. The type of segmentation task varies depending on the definition of the concepts, such as foreground, category, or object instance. For example, semantic segmentation~\cite{zhou2018ade} involves pixel-level semantic classification of an image, while instance segmentation~\cite{lin2014coco} aims to identify different object instances and their categories. 
Video object segmentation~\cite{xu2018youtube,pont20172017,ding2023mose} is the task of segmenting a particular object throughout the entire video sequence given only the object mask of the first frame.

Previous segmentation methods~\cite{long2015fully,lin2017refinenet,pspnet,ranftl2021vision,he2017mask,de2017semantic,yolact,wang2021SOLO,panopticfpn,carion2020detr,mask2former} have been designed specifically for certain tasks and cannot be generalized for switching tasks or changing categories. This paper introduces a general interface that is compatible with all segmentation tasks with an appropriate training scheme, a single generalist model can achieve good performance on both in-domain and out-of-domain segmentation tasks, either qualitatively or
quantitatively.

\subsection{Vision Generalist}
In recent years, there have been efforts to unify different tasks in the vision domain using Transformer-based models, resulting in several vision generalists~\cite{Chen2021pix2seq,Chen2022pix2seq2,Zhu2022uniperceiver,Lu2022unifiedio,Kolesnikov2022UVIM}.
DETR~\cite{carion2020detr} is one of the first to adopt Transformer~\cite{vaswani2017attention} as a task-specific head for object detection. 
Pix2Seq series~\cite{Chen2021pix2seq,Chen2022pix2seq2} defines the output spaces of vision tasks as discrete ones and performs the task of object detection, instance segmentation, keypoint estimation, and image captioning, in an auto-regressive manner. 
Unified-IO~\cite{Lu2022unifiedio} and OFA~\cite{Wang2022OFA} perform joint modeling across vision, vision \& language, and NLP tasks in a sequence-to-sequence manner, that both the inputs and outputs are defined to a sequence of discrete tokens. 
UViM~\cite{Kolesnikov2022UVIM} unifies pixel-labeling tasks together, such as panoptic segmentation, depth estimation, and colorization, but trains separate models for each.

Although these works all appear to unify different tasks into similar spaces, they actually accomplish each task through some form of hard indicators, such as a special token, making it difficult to generalize to new tasks. In contrast, this work uses an in-context framework that maintains flexibility on task definition and utilizes a random coloring scheme to prevent the model from collapsing into a multi-task learning solution and instead forces it to accomplish the assigned task via referring contextual information. 
Another difference is the scope of the tasks. This work primarily focuses on a crucial category in visual perception, namely image segmentation.

\subsection{In-Context Visual Learning}
GPT-3~\cite{gpt3} introduces the concept of in-context learning to deep learning, which allows a series of NLP tasks to be formulated as text completion problems given prompts and examples. In computer vision, \cite{Bar2022VisualPrompt}~first proposes an in-context training framework using inpainting with discrete tokens on figures and infographics from vision articles, demonstrating the framework's capabilities in foreground segmentation, single object detection, and colorization. Painter~\cite{painter} adopts masked image modeling on continuous pixels to perform in-context training with supervised datasets, on seven diverse and challenging vision tasks, achieving highly competitive results on these tasks.

Our work builds upon the Painter framework, but with a specific focus on the segmentation task due to its central role in visual perception. Thus this work unifies diverse segmentation data including semantic segmentation, instance segmentation, part segmentation, and even those for special scenarios like aerial images. 
Additionally, we design a random coloring scheme that forces the model to reference contextual information to complete the assigned task but not collapse into the multi-task solution. As segmentation tasks and datasets have less variability than depth/pose estimation, it is easier to share internal structures for effective training of in-domain tasks, while maintaining the generalization capability to out-of-domain segmentation tasks.

\section{Approach}

\Ours is a special version of Painter~\cite{painter} framework which enables to \textbf{seg}ment everything with a \textbf{g}eneralist \textbf{P}ain\textbf{t}er, thus the name of our model, \textbf{\Ours}.
This training framework redefines the output space of vision tasks as ``images'' and unifies different tasks into the same image inpainting problem, \ie, to randomly mask the task output images and reconstruct the missing pixels.
To maintain the simplicity and generality, we make no modifications to the architecture and loss function, \ie, a vanilla ViT~\cite{dosovitskiy2020vit} and a simple smooth-$\ell_1$~\cite{girshick2015fast} loss, but design a new random coloring scheme in in-context training for better generalization capability.

\subsection{In-Context Coloring}

In the traditional framework of Painter, the color space for each task is pre-defined, resulting in the solution collapse into multi-task learning.
For example, for semantic segmentation, a set of colors is pre-defined, and each semantic category is assigned a fixed color.
Similarly, in instance segmentation, the color of an instance object is assigned according to its location categories, \ie, the number of colors equals the number of spatial locations, resulting in the model only relying on the color itself to determine the task, rather than using the relationships between segments.

To address this limitation, we propose a random coloring scheme for in-context coloring. We begin by randomly sampling another image that shares a similar context with the input image, such as the same semantic category or object instance. Next, we randomly sample a set of colors from the target image and map each color to a random one. This results in a re-coloring of the corresponding pixels. 
As a result, we get two pairs of images, which are defined as an in-context pair.
In addition, we introduce the mix-context training method which trains the model using mixed examples. This involves stitching together multiple images with the same color mapping. The resulting image is then randomly cropped and resized to form a mixed-context training sample. 
By doing so, the model learns to focus on the contextual information of the image rather than just relying on specific color information to determine the task.

Such unification allows us to utilize all segmentation datasets in a consistent way, only varying the data sampling strategy depending on the specific task.
We define different contexts according to different data types.
For semantic segmentation, we randomly sample the categories.
For instance segmentation, object instances are sampled in random numbers.
The different views of the same image, \eg, transformed by a set of augmentations, are treated as the images in context.
In the implementation, the sampling is all about colors, \eg, the same color refers to either the same category or the same instance.

\begin{figure}
    \centering
    \includegraphics[width=0.95\linewidth]{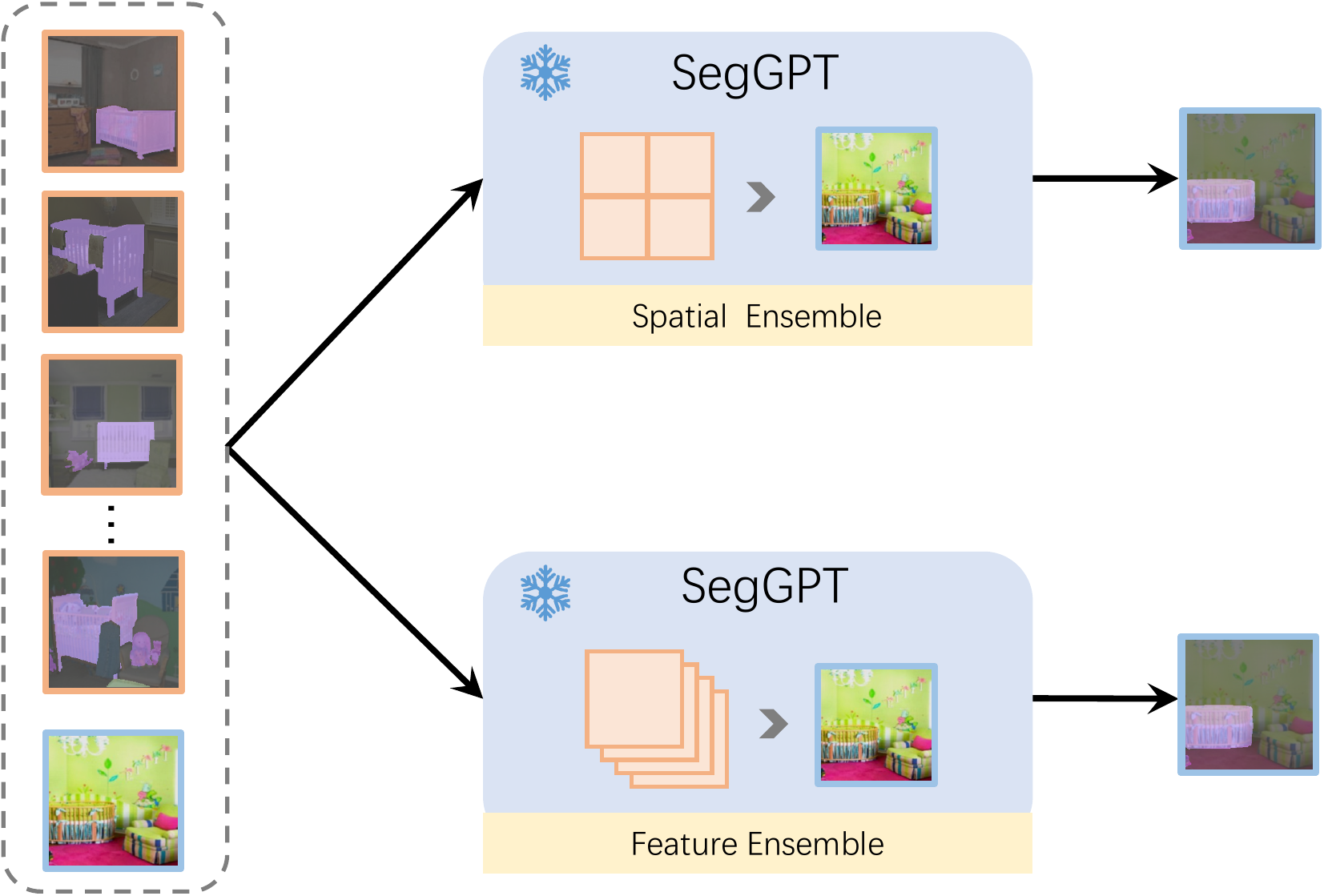}
    \caption{Illustration of our proposed context ensemble strategies for multi-example inference: the spatial ensemble (top) and the feature ensemble (bottom). The spatial ensemble strategy involves stitching multiple example images together and resizing them to the input resolution. The feature ensemble strategy averages features of the query image after each attention layer so that the query image aggregates all the reference examples.}
    \label{fig:method_ensemble}
\end{figure}

\subsection{Context Ensemble}
\label{sec:ensemble}

Once the training is finished, its full power can be unleashed during inference.
\Ours enables arbitrary segmentation in context, \eg, with an example of a single image and its target image.
The target image can be of a single color (excluding the background), or multiple colors, \eg, segmenting several categories or objects of interest in one shot.
Specifically, given an input image to be tested, we stitch it with the example image and feed it to \Ours to get the corresponding in-context predictions.

To serve a more accurate and concrete context, multiple examples can be used.
For instance, several examples of the same semantic category, or the previous frames in a video, can be employed.
To efficiently leverage multiple examples for a \Ours model, we propose two context ensemble approaches.
One is called $\tt Spatial$ $\tt Ensemble$, multiple examples concatenated in $n \times n$ grid and then sub-sampled to the same size as a single example. This approach is in line with the intuition of in-context coloring and the semantic information of multiple examples can be in-context extracted with almost no additional cost. 
Another approach is $\tt Feature$ $\tt Ensemble$. Multiple examples are combined in the batch dimension and computed independently except that features of the query image are averaged after each attention layer. In this way, the query image gathers information about multiple examples during inference.

\begin{figure}
    \centering
    \includegraphics[width=0.97\linewidth]{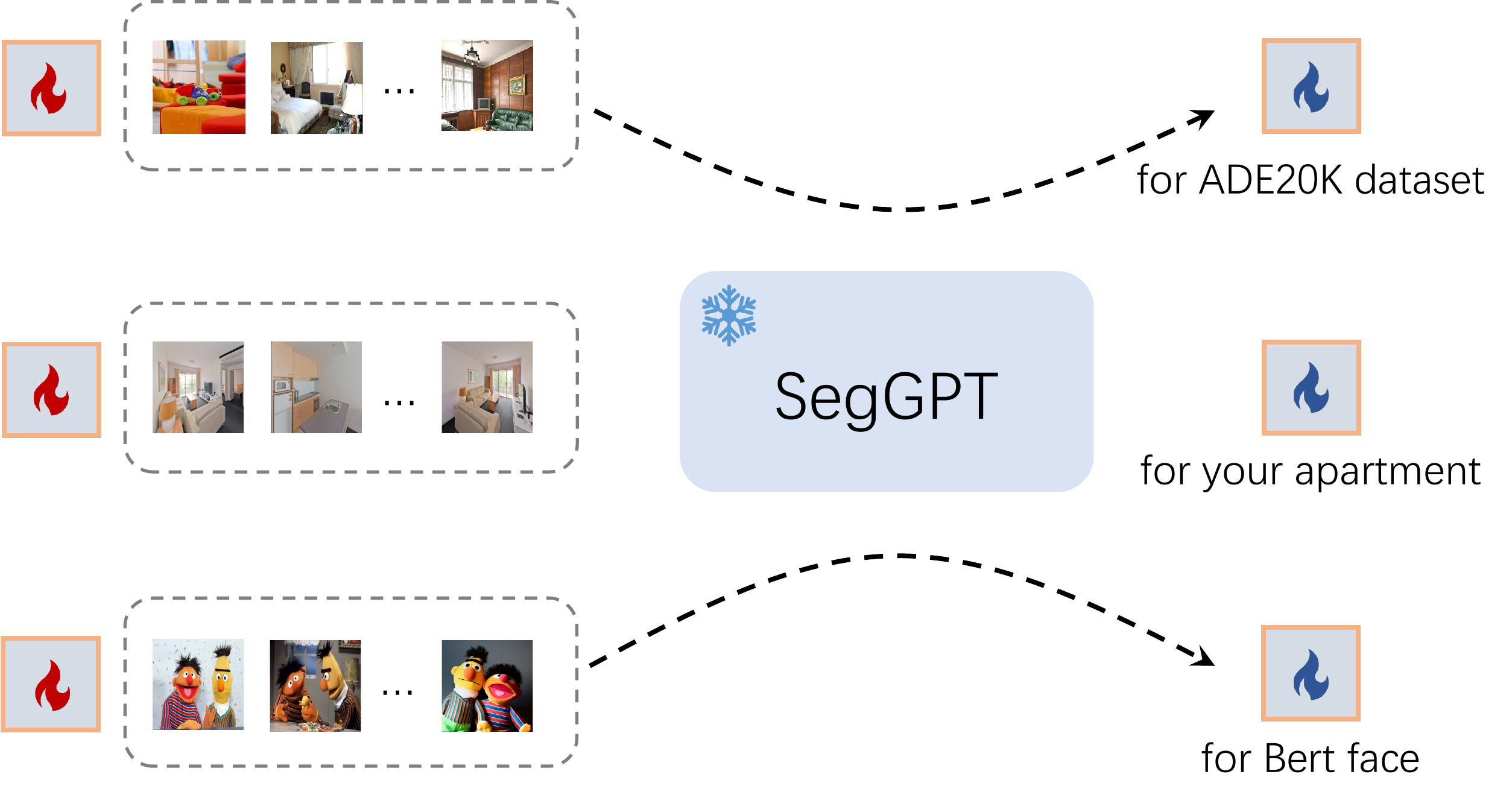}
    \caption{Illustration of in-context tuning on different task specifications. For in-context tuning, we freeze the whole pre-trained model and only optimize the learnable image tensor which serves as the input context. We can perform the in-context prompt tuning on the specific datasets (ADE-20K semantic segmentation), specific scenes (your apartment), and even specific characters (Bert's face). }
    \label{fig:method_tuning}
\end{figure}

\begin{figure*}
    \centering
    \includegraphics[width=0.98\linewidth]{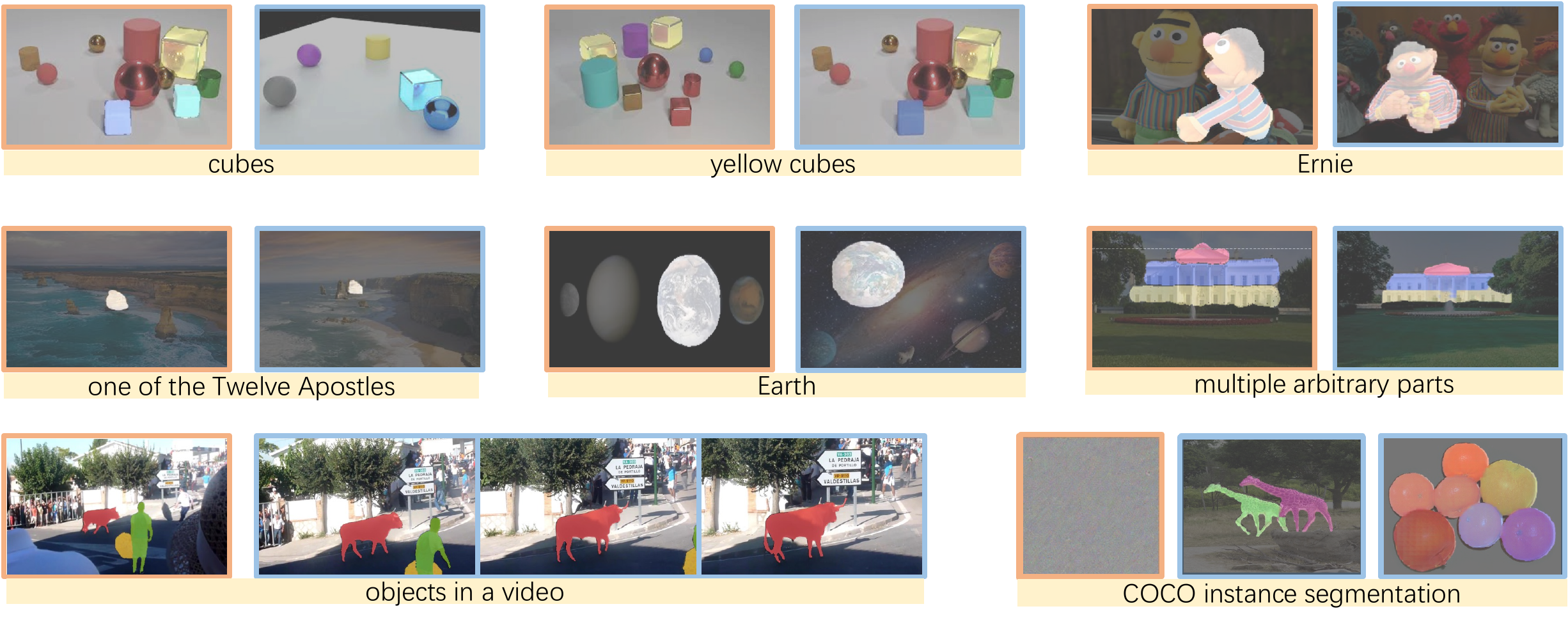}
    \caption{More visualizations. For each sample, the orange box \textcolor{orange}{$\square$} on the left displays the example/prompt image and its corresponding mask, while the blue box \textcolor{cyan}{$\square$} on the right shows the input image and the resulting mask output. The mask is visualized via the bright region attached to the image. \Ours can perform arbitrary object/part segmentation (cubes, yellow cubes, Ernie, one of the Twelve Apostles, earth, multiple arbitrary parts), video object segmentation without videos in training, and close-set instance segmentation on COCO with learnable prompt tuning.}
    \label{fig:vis_2}
\end{figure*}

\subsection{In-Context Tuning}

\Ours is capable of adapting to a unique use case without updating the model parameters.
We freeze the whole model and initialize a learnable image tensor as the input context.
Only this learnable image tensor is updated during the training.
The rest of the training remains the same, \eg, the same loss function.
After the tuning, we take the learned image tensor out and use it as a plug-and-play key for a specific application.
For example, given a dataset with a fixed set of object categories, \eg, ADE20K, we could train a customized prompt for this dataset, while there is no harm to the generality of the model.
Or, we could optimize a prompt image for a specific scene, \eg, your apartment, or a specific character, \eg, Bert's face.
This opens up opportunities for a broad range of applications.

\section{Experiment}

\begin{figure*}
    \centering
    \includegraphics[width=0.92\linewidth]{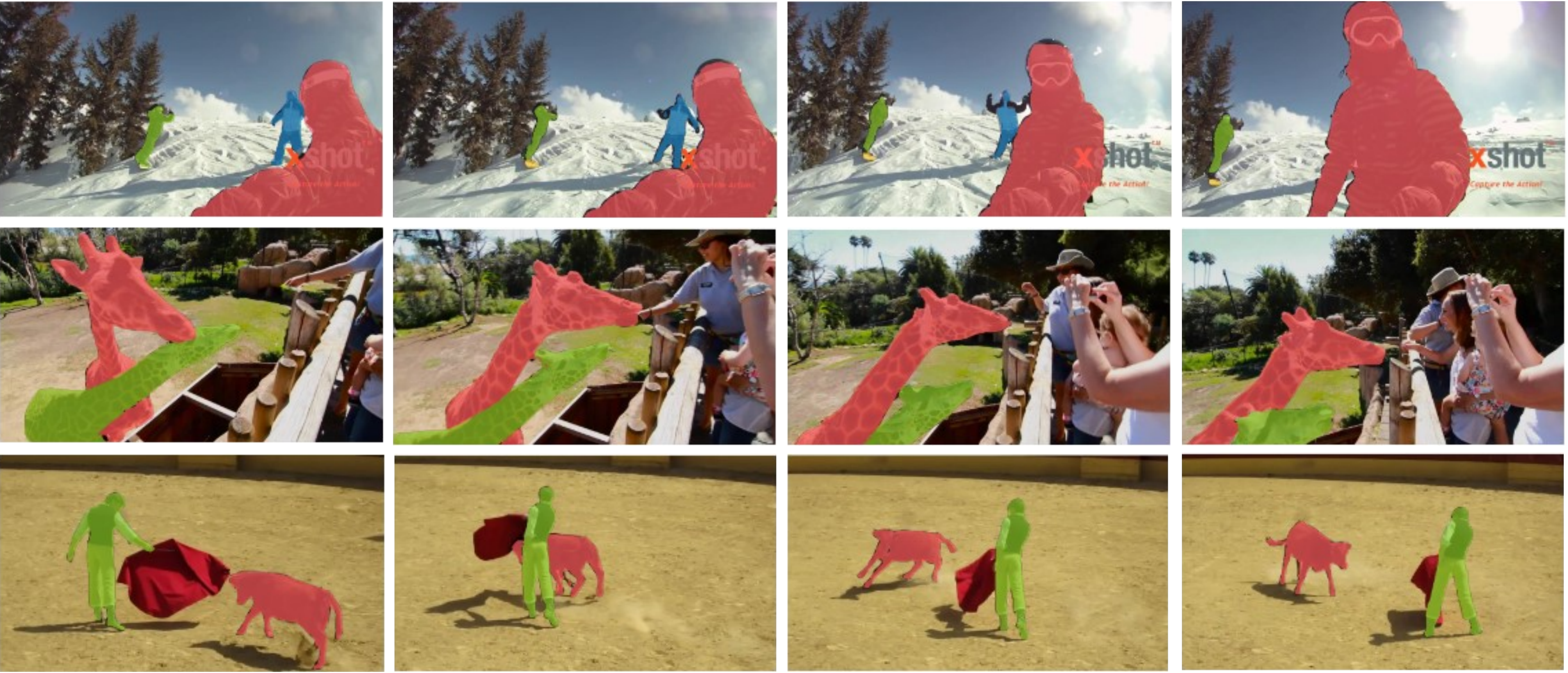}
    \caption{Qualitative results of video object segmentation on YouTube-VOS 2018.}
    \label{fig:vosvis}
\end{figure*}

\subsection{Training Data}

Our approach uses a diverse set of segmentation datasets, including part, semantic, instance, panoptic, person, retinal-vessel, and aerial-image segmentation. Unlike previous methods that relied on handcrafted label merging to combine different types of segmentation datasets, our method offers a unified perspective that eliminates the need for additional effort or adjustment on the datasets. In particular, our approach does not require any modifications to either the architecture or training pipeline when adding an extra dataset.

\myparagraph{ADE20K}~\cite{zhou2018ade}
 provides segmentation labels for 150 semantic categories, with a total of 
25K images, including 20K training images, 2K validation images, and  3K  testing images.

\myparagraph{COCO}~\cite{lin2014coco}
 is a widely used visual perception dataset that supports instance segmentation, semantic segmentation and panoptic segmentation.
It contains 118K training images and 5K validation, with 80 ``things'' and 53 ``stuff'' categories.

\myparagraph{PASCAL VOC}~\cite{everingham2009pascalvoc}
is a classic object recognition dataset.
We use the augmented segmentation version which provides annotations of 20 categories on 10582 training images.

\myparagraph{Cityscapes}~\cite{cordts2016cityscapes}
 focuses on the scene understanding of the street views.
We use the 2954 training images with semantic segmentation annotations of 19 categories.

\myparagraph{LIP}~\cite{LIP_dataset}
 focuses on the semantic understanding of the person. We use the 30385 training images with segmentation labels of 19 human part categories.

\myparagraph{PACO}~\cite{ramanathan2023paco}
is a newly released dataset that provides annotations for the parts and attributes of  common objects.
We process and use the 41807 training images with part annotations.

\myparagraph{CHASE\_DB1}~\cite{chase_db1_data}, \textbf{DRIVE}~\cite{drive_dataset}, \textbf{HRF}~\cite{HRF_dataset} and \textbf{STARE}~\cite{stare_dataset}
provide annotations for retinal vessel segmentation. We augment the high-resolution raw images with random cropping.

\myparagraph{iSAID}~\cite{waqas2019isaid_dataset} and \textbf{loveDA~\cite{wang2021loveda_dataset}}
focus on semantic understanding in aerial images, with 23262 and 2520 training images for 15 and 6 semantic categories respectively.

\begin{table*}[ht]
\centering
\small 
\begin{tabular}{ll|cc|cc}
\multirow{2}{*}{method} & \multirow{2}{*}{venue}
& \multicolumn{2}{c|}{COCO-20$^i$} & \multicolumn{2}{c}{PASCAL-5$^i$}            \\
&
& one-shot     & few-shot       & one-shot     & few-shot                  \\
\shline
\textit{\small specialist model} &&&&&\\
\gr{HSNet~\cite{min2021hypercorrelation}}           & \multirow{2}{*}{ICCV'21}
& \gr{41.2}         & \gr{49.5}           & \gr{66.2}         & \gr{70.4}                      \\
HSNet*                                         &
& 41.7         & 50.7           & 68.7         & 73.8                      \\
\gr{VAT~\cite{hong2022cost}}                        & \multirow{2}{*}{ECCV'22}
& \gr{41.3}         & \gr{47.9}           & \gr{67.9}         & \gr{72.0}                      \\
VAT*                                           & 
& 42.9         & 49.4           & 72.4         & 76.3                      \\
\gr{FPTrans~\cite{zhang2022FPTrans}}                & \multirow{2}{*}{NeurIPS'22}
& \gr{47.0}         & \gr{58.9}           & \gr{68.8}         & \gr{78.0}                      \\
FPTrans*                                       &
& 56.5         & 65.5           & 77.7         & 83.2                      \\
\shline
\textit{\small generalist model} &&&&&\\
Painter                                        & CVPR'23
& 32.8         & 32.6           & 64.5        & 64.6                      \\
\Ours                                          & this work
& 56.1         & 67.9          & 83.2         & 89.8                      \\

\end{tabular}
\caption{Quantitative results on COCO-20$^i$ and PASCAL-5$^i$ of example-based semantic segmentation. * indicates that the categories in training cover the categories in testing. }
\label{table:fewshot_sem_seg}
\end{table*}

\subsection{One-Shot Training Details}
Our approach for segmentation tasks utilizes a general interface, where we emphasize that we only train one generalist model with a mixture of datasets, and evaluated this model on diverse benchmarks. 
Following~\cite{painter}, we use a Vision Transformer (ViT-L) encoder~\cite{dosovitskiy2020vit}, which has 307M parameters.
We use a pre-trained checkpoint from~\cite{painter} as the initialization.
We employ an AdamW optimizer~\cite{kingma2014adam} and a \emph{cosine} learning rate scheduler, with a base learning rate 1e$-4$.
Weight decay is set to 0.05. The batch size is 2048. We train for 9K iterations, with a warm-up period of 1.8K iterations.
We use a set of data augmentations including random resize cropping, color jittering, and random horizontal flipping. The size of a single input image is $448\times 448$.

\subsection{Qualitative Results}
To demonstrate the capability of our \Ours in an intuitive perspective, we visualize the task output of the selected images with the specialized task prompts, shown in Figure~\ref{fig:teaser} and Figure~\ref{fig:vis_2}. 
These two figures include a wide range of segmentation tasks, such as arbitrary part/object segmentation with varied granularities, text segmentation, video object segmentation without videos in training, and close-set instance/semantic segmentation with learnable prompt tuning.
Figure~\ref{fig:vosvis} presents more visualizations on video object segmentation of YouTube-VOS 2018 dataset.
From these visualizations, \Ours demonstrates the ability to make highly accurate predictions across a wide range of tasks, while maintaining super flexibility in the task definition.

\begin{table}[ht]
\centering
\begin{tabular}{ll|cc}
\multirow{2}{*}{method} & \multirow{2}{*}{venue} & \multicolumn{2}{c}{mIoU}  \\
                        &                              & one-shot  & few-shot      \\
\shline
\multicolumn{2}{l|}{\textit{\small trained on FSS-1000}} \\
\gr{DAN~\cite{wang2020few}}                   & \gr{ECCV'20}
& \gr{85.2}        & \gr{88.1}                                \\
\gr{HSNet~\cite{min2021hypercorrelation}}     & \gr{ICCV'21}
& \gr{86.5}        & \gr{88.5}                       \\
\gr{SSP~\cite{fan2022self}}                   & \gr{ECCV'22}
& \gr{87.3}        & \gr{88.6}                    \\
\gr{VAT~\cite{hong2022cost}}                  & \gr{ECCV'22}
& \gr{90.3}        & \gr{90.8}                    \\
\gr{DACM~\cite{xiong2022doubly}}              & \gr{ECCV'22}
& \gr{90.8}        & \gr{91.7}                    \\
\shline
\multicolumn{2}{l|}{\textit{\small not trained on FSS-1000}} \\
Painter                                  & CVPR'23
& 61.7             & 62.3                    \\
\Ours                                    & this work %
& 85.6             & 89.3                    \\
\end{tabular}
\caption{Quantitative results on few-shot semantic segmentation on FSS-1000. \Ours achieves remarkable results although not trained on FSS-1000.}
\label{table:fss1000_sem_seg}
\end{table}

\begin{table*}[ht]
\centering
\small 
\begin{tabular}{ll|ccccc|ccc|ccc}
\multirow{2}{*}{method} & \multirow{2}{*}{venue}
& \multicolumn{5}{c|}{YouTube-VOS 2018~\cite{xu2018youtube}}  & \multicolumn{3}{c|}{DAVIS 2017~\cite{pont20172017}} & \multicolumn{3}{c}{MOSE~\cite{ding2023mose}}\\
&
& $G$ & $J_s$ & $F_s$ & $J_u$ & $F_u$   & $J\&F$ & $J$ & $F$ & $J\&F$ & $J$ & $F$  \\
\shline
\textit{\small with video data} &&&&&&&&&&&\\
\gr{AGAME~\cite{johnander2019generative}}      & \gr{CVPR'19}
& \gr{66.0} & \gr{66.9} & \gr{-}    & \gr{61.2} & \gr{-}    & \gr{70.0} & \gr{67.2} & \gr{72.7} & \gr{-}    & \gr{-}    & \gr{-}      \\
\gr{AGSS~\cite{lin2019agss}}                   & \gr{ICCV'19}
& \gr{71.3} & \gr{71.3} & \gr{65.5} & \gr{75.2} & \gr{73.1} & \gr{67.4} & \gr{64.9} & \gr{69.9} & \gr{-}    & \gr{-}    & \gr{-}      \\
\gr{STM~\cite{oh2019video}}                    & \gr{ICCV'19}
& \gr{79.4} & \gr{79.7} & \gr{84.2} & \gr{72.8} & \gr{80.9} & \gr{81.8} & \gr{79.2} & \gr{84.3} & \gr{-}    & \gr{-}    & \gr{-}      \\
\gr{AFB-URR~\cite{liang2020video}}             & \gr{NeurIPS'20}
& \gr{79.6} & \gr{78.8} & \gr{83.1} & \gr{74.1} & \gr{82.6} & \gr{74.6} & \gr{73.0} & \gr{76.1} & \gr{-}    & \gr{-}    & \gr{-}      \\
\gr{RDE~\cite{li2022recurrent}}                & \gr{CVPR'22}
& \gr{83.3} & \gr{81.9} & \gr{86.3} & \gr{78.0} & \gr{86.9} & \gr{86.1} & \gr{82.1} & \gr{90.0} & \gr{48.8} & \gr{44.6} & \gr{52.9}   \\
\gr{SWEM~\cite{lin2022swem}}                   & \gr{CVPR'22}
& \gr{82.8} & \gr{82.4} & \gr{86.9} & \gr{77.1} & \gr{85.0} & \gr{84.3} & \gr{81.2} & \gr{87.4} & \gr{50.9} & \gr{46.8} & \gr{54.9}   \\
\gr{XMem~\cite{cheng2022xmem}}                 & \gr{ECCV'22}
& \gr{86.1} & \gr{85.1} & \gr{89.8} & \gr{80.3} & \gr{89.2} & \gr{87.7} & \gr{84.0} & \gr{91.4} & \gr{57.6} & \gr{53.3} & \gr{62.0}   \\
\shline
\textit{\small without video data}  &&&&&&&&&&&\\
Painter                                        & CVPR'23
& 24.1 & 27.6 & 35.8 & 14.3 & 18.7 & 34.6 & 28.5 & 40.8 & 14.5 & 10.4 & 18.5   \\
\Ours                                          & this work %
& 74.7 & 75.1 & 80.2 & 67.4 & 75.9 & 75.6 &  72.5 & 78.6 & 45.1 & 42.2 & 48.0  \\
\end{tabular}
\caption{Quantitative results of video object segmentation on YouTube-VOS 2018, DAVIS 2017, and MOSE. Notably, Painter and \Ours do not use any video data in training. $G$ is the average score over ``seen" and ``unseen" classes in YouTube-VOS 2018.}
\label{table:vos}
\end{table*}

\subsection{Comparison with Specialist Methods}

\myparagraph{Few-shot semantic segmentation.} We evaluate the performance of \Ours, on two settings of few-shot semantic segmentation: in-domain on COCO-20$^i$/PASCAL-5$^i$, and out-of-domain on FSS-1000. 
Table~\ref{table:fewshot_sem_seg} shows the results of example-based semantic segmentation on COCO-20$^i$/PASCAL-5$^i$. For a fair comparison, we also evaluate specialist models on in-domain categories marked by *. Our results indicate that \Ours can achieve comparable or significantly better performance than recently published state-of-the-art specialist models on these two benchmarks. 
Note that the prior art FPTrans trains separate models with different shots.
Furthermore, \Ours surpasses the generalist Painter~\cite{painter} by a considerable margin.

Table~\ref{table:fss1000_sem_seg} presents the results of few-shot semantic segmentation on FSS-1000 with out-of-domain categories.
Compared to specialist models trained on FSS-1000, \Ours exhibits highly competitive performance. Notably, our model is not trained on the FSS-1000 dataset at all, yet still achieves remarkable results, demonstrating its effectiveness.

\myparagraph{Video object segmentation.}
Video object segmentation (VOS) is a task that segments a particular object in video frames. In this work, we focus on the semi-supervised VOS setting and evaluate our proposed method, \Ours, on the validation split of three datasets: YouTube-VOS 2018~\cite{xu2018youtube}, DAVIS 2017~\cite{pont20172017}, and the recently release challenging benchmark MOSE~\cite{ding2023mose}.
We use two metrics commonly used in VOS for evaluation: the $J$ score and the $F$ score, and we evaluate our results with official evaluation servers or tools. 

\Ours performs video object segmentation by converting the first frame and its object mask to in-context coloring examples. When testing a current frame, we use its previous $K$ frames (if have) for constructing multiple examples. Object masks for these frames have been predicted and stored by a FIFO queue. After multiple examples are constructed, \texttt{Feature Ensemble} (describe in Section~\ref{sec:ensemble}) is applied and the prediction result will be stored for the next frame. 
We evaluate our model on several benchmarks, and the results are presented in Table~\ref{table:vos}. Despite not being specifically trained for the task, our approach achieves competitive results with the specialist models trained on these datasets. For instance, on YouTube-VOS 2018~\cite{xu2018youtube}, our method outperformed the task-specific approach AGAME~\cite{johnander2019generative} and AGSS~\cite{lin2019agss} by clear margins.
On the challenging MOSE benchmark which focuses on complex scenes, \Ours even performs comparably with the state-of-the-art method RDE.

\begin{table*}[ht]
\centering
\begin{subtable}[t]{0.55\linewidth}
\centering
\begin{tabular}{cl|ccc|cc}
\multirow{2}{*}{examples} & \multirow{2}{*}{ensemble}
& \multicolumn{3}{c|}{DAVIS 2017}       & \multicolumn{2}{c}{FSS-1000} \\
&
& $J\&F$     & $J$      & $F$        & mIoU         & FB-IoU        \\
\shline
1        & -
& 70.0       & 66.4     & 73.7       & 85.5         & 90.8          \\
4        & Spatial
& 61.9       & 58.0     & 65.8       & 89.3         & 93.5          \\
4        & Feature
& 74.7       & 71.6     & 77.7       & 87.8         & 92.4          \\
8        & Feature
& 75.6       & 72.5     & 78.6       & 89.8         & 93.8          \\
\end{tabular}
\caption{}
\label{table:ensemble}
\end{subtable}
\begin{subtable}[t]{0.44\linewidth}
\centering
\begin{tabular}{c|ccccc}
& \multicolumn{5}{c}{DAVIS 2017}            \\
frames
& 1     & 4     & 8     & 12     & 16     \\
\shline
$J\&F$
& 70.0  & 74.7  & 75.6  & 74.8   & 74.6   \\
$J$
& 66.4  & 71.6  & 72.5  & 71.6   & 71.4   \\
$F$
& 73.7  & 77.7  & 78.6  & 77.9   & 77.8   \\
\end{tabular}
\vspace{6pt}
\caption{}
\label{table:nframes}
\end{subtable}
\caption{Ablation study on ensemble strategy (a) and the number of frames (b) in in-context inference. Spatial ensemble approach performs well on FSS-1000 dataset but experiences a performance drop on DAVIS 2017. Feature ensemble achieves better results due to no sub-sampling.}
\end{table*}

\subsection{Ablation Study}

Here we ablate two context ensemble strategies, namely spatial and feature ensemble. Results are shown in Table~\ref{table:ensemble}. Our findings reveal that the spatial ensemble approach performs well on FSS-1000 dataset but experiences a performance drop on DAVIS 2017. We attribute this to the fact that the spatial ensemble employs the sub-sampling on the examples. Notably, FSS-1000 dataset has a lower image resolution (224$\times$224) compared to the high-resolution DAVIS dataset (640$\times$480), and therefore, sub-sampling does not result in significant information loss for FSS-1000.
While, we observe that feature ensemble can reduce this information loss on sub-sampling, and achieve significantly better performance on DAVIS 2017.

We also ablate the number of frames in DAVIS 2017, as shown in Table~\ref{table:nframes}. As the number of frames increases, the performance initially improves before reaching a point of diminishing returns. In particular, we observe that the optimal performance is achieved when using 8 frames.

\begin{table}[t]
\centering
\small 
\begin{tabular}{ll|c}
 method & venue & mIoU  \\
\shline
\textit{\small specialist model} &&\\
\gr{FCN~\cite{long2015fully}} & \gr{CVPR'15} & \gr{29.4} \\
\gr{RefineNet~\cite{lin2017refinenet}} & \gr{CVPR'17} & \gr{40.7} \\
\gr{DPT~\cite{ranftl2021vision}} & \gr{ICCV'21} & \gr{49.2} \\
\gr{Mask2Former~\cite{mask2former}} & \gr{CVPR'22} &  \gr{57.7} \\
\shline
\textit{\small generalist model} &&\\
Painter & CVPR'23 & 49.9 \\
\Ours & %
this work 
& 39.6 \\
\end{tabular}
\caption{Results on ADE20K semantic segmentation. }%
\label{table:ade20k}
\end{table}

\subsection{In-Context Tuning}
In-context tuning enables to customize a unique application with a set of data samples.
For example, to tune a prompt for a specific dataset, scene, or even a person.
Specifically, we define the task prompt as the learnable tensors, freeze the whole model, and then use the same training loss to optimize the task prompts.
Here, we conduct in-context tuning on the challenging ADE20K semantic segmentation and COCO panoptic segmentation. We evaluate \Ours with learnable prompts on the corresponding benchmarks.

Results on ADE20K semantic segmentation are shown in Table~\ref{table:ade20k}. Our model \Ours achieves competitive performance with specialist models like RefineNet. 
However, compared to the generalist Painter, our approach shows a 10.3 point drop in mIoU. 
This observation can be explained by the introduction of a random color scheme, which makes it more challenging for the model to use color as a simple indicator of in-domain tasks. Instead, the model needs to rely on context examples to determine the task, making optimization much more difficult.
Similarly, Table~\ref{table:coco_pano} shows the results of our \Ours model on COCO panoptic segmentation. Here, we again observe a 9.0 point drop in PQ compared to the generalist Painter.
Outperforming all specialist methods in specific benchmarks is not the purpose of this work, and we believe there is much room to improve in the future.

\begin{table}[ht]
\centering
\small 
\begin{tabular}{ll|c}
 method & venue & PQ \\
\shline
\textit{\small specialist model} &&\\
\gr{PanopticFPN~\cite{panopticfpn}} & \gr{CVPR'19} & \gr{40.3} \\
\gr{SOLOv2~\cite{wang2020solov2}} & \gr{NeurIPS'20} &  \gr{42.1}  \\
\gr{Mask2Former~\cite{mask2former}} & \gr{CVPR'22} & \gr{57.8}  \\
\gr{UViM~\cite{Kolesnikov2022UVIM}} & \gr{NeurIPS'22} & \gr{45.8} \\
\shline
\textit{\small generalist model} &&\\
Painter & CVPR'23 & 43.4  \\
\Ours   &  
this work 
& 34.4  \\
\end{tabular}
\caption{Results on COCO panoptic segmentation. }%
\label{table:coco_pano}
\end{table}

\section{Discussion and Conclusion}

In this work, we present a generalist segmentation model, showing how to design an appropriate training strategy to fully leverage the flexibility of in-context visual learning. Our model exhibits strong capabilities in handling both in-domain and out-of-domain segmentation tasks, including object instance, stuff, part, contour, text segmentation, \etc.

This work is not without drawbacks. While our work introduces a new random coloring regime for better generalization capability of in-context training, it also makes the training task inherently more difficult, which may be the reason for inferior performance in in-domain tasks with ample training data, such as semantic segmentation on ADE20K and panoptic segmentation on COCO. %

Looking forward, we believe that our approach has the potential to serve as a powerful tool for enabling more diverse applications in image/video segmentation, by leveraging the flexibility in task definition 
with 
in-context inference.
Scaling up model size is one avenue that we plan to pursue to further improve performance. With larger models, more complex patterns in the data can be captured, which may lead to better segmentation results. However, this comes with the challenge of finding more data. One potential solution is to explore self-supervised learning techniques.
We hope that our work will inspire the community to continue exploring the potential of in-context learning in computer vision. 
We 
remain optimistic that the best GPT-3 moment in the vision field is yet to come.

\section*{Acknowledgement}
\label{sec:ack}
This project is supported by the National Key R\&D Program of China (2022ZD0116302).
We would like to thank Yemin Shi and Teng Dai for their help on the demo, Hanxiao Qu, Yan Tian, and Xigang Cao for the help on GPU resources, as well as other colleagues at Beijing Academy of Artificial Intelligence for support throughout this project.

{\small
\bibliographystyle{ieee_fullname}
\bibliography{egbib}
}

\appendix

\section*{Appendix}

\renewcommand{\thefigure}{S\arabic{figure}}
\setcounter{figure}{0}
\renewcommand{\thetable}{S\arabic{table}}
\setcounter{table}{0}

\section{Additional Implementation Details}

\myparagraph{Training.}
We use various segmentation datasets during training. The sampling weight for each dataset is 0.22 (COCO  instance), 0.15 (ADE20K semantic), 0.15 (COCO panoptic semantic), 0.07 (Cityscapes semantic), 0.07 (COCO stuff semantic), 0.07 (LIP person semantic), 0.07 (PASCAL VOC semantic), 0.07 (PACO semantic), 0.06 (iSAID and loveDA aerial semantic), and 0.06 (CHASE\_DB, DRIVE, HRF and STARE retinal vessel).
For semantic segmentation data, we use a probability of 0.5 for using the transformation of the input image  as the in-context examples and then conduct random color selection. 
For instance segmentation data, the probability is 1.0, \ie, we always use two transformed views of the same image as the in-context pair.
Almost all the segmentation sub-tasks can be grouped into two types, \ie, to segment a category or an instance (not limited to objects).
To avoid the ambiguity between category and instance,  we initialize two learnable embeddings which are associated with category-level and instance-level coloring tasks respectively.

\myparagraph{Evaluation.}
For quantitative evaluation on the existing benchmarks, the examples are either from the support samples, the training set, the first frame in a video, or a learned prompt.
Take ADE20K semantic segmentation as an example.
Given a tuned prompt, we directly stitch the prompt with each test image to obtain the predictions.
Without the tuned prompt, for each category, we randomly sample several images from the training set which contain that category.
These examples are used together via context ensemble to obtain the predictions for this category across all test images.

\section{Additional Results}
\myparagraph{ADE20K semantic segmentation.}
In Table~\ref{table:supp_ade20k}, we provide the example-based semantic segmentation results on ADE20K. Different from the in-context tuning, we only randomly select several samples in the training set as examples, and use \texttt{Feature Ensemble} to ensemble the examples.
Specifically, for each category, we randomly sample without replacement from all images with that category. Since the selection of the examples can affect performance, we sample with different random seeds \{$1000$, $2000$, $3000$, $4000$\} and report the best results. 
We can see that more examples significantly boost the performance, \eg, +13.1\% mIoU from 1 to 16 examples, although there is still a gap with the  tuned prompt.
These experiments inspire us to explore in the future what makes good examples and how many examples we need to approach the results of in-context tuning.

\begin{figure}[th]
    \centering
    \vspace{-1.2pt}
    \begin{subfigure}{1.0\linewidth}
        \centering
        \setlength{\abovecaptionskip}{0.5pt}
        \includegraphics[width=0.9\linewidth, trim=200 0 0 0, clip]{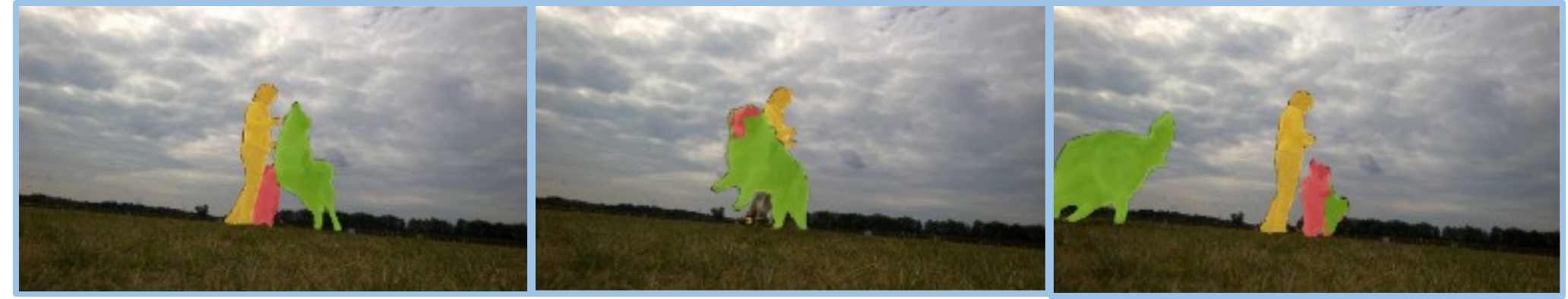}
        \caption{}
        \vspace{4pt}
    \end{subfigure}
    \begin{subfigure}{1.0\linewidth}
        \centering
        \setlength{\abovecaptionskip}{0.5pt}
        \includegraphics[width=0.9\linewidth, trim=200 0 0 0, clip]{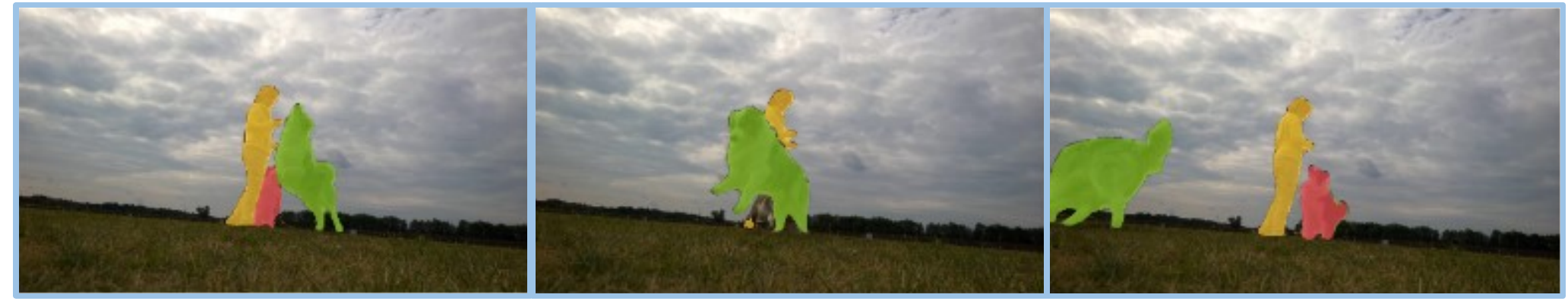}
        \caption{}
        \vspace{-0.5em}
    \end{subfigure}
    \vspace{-1.0em}
    \caption{Context ensemble helps segment objects across frames. a) Incorrect predictions for objects in a crowd when only the first frame is used as the example. b) Correct predictions using \texttt{Feature Ensemble} with previous frames.}
    \label{fig:demonstration}
    \vspace{-8pt}
\end{figure}

\myparagraph{Context ensemble.}
Here we qualitatively demonstrate the effectiveness of our context ensemble approach in Figure~\ref{fig:demonstration}.
Given a video clip and its first annotated frame, it is difficult to distinguish the instances in a crowd when using only the first frame as an example.
With the context ensemble of several previous frames and their pseudo-labels, \Ours segments each object successfully.

\begin{figure*}[!h]
    \centering
    \begin{subfigure}{0.85\linewidth}
        \centering
        \setlength{\abovecaptionskip}{0.5pt}
        \includegraphics[width=0.99\linewidth, trim=0 0 0 0, clip]{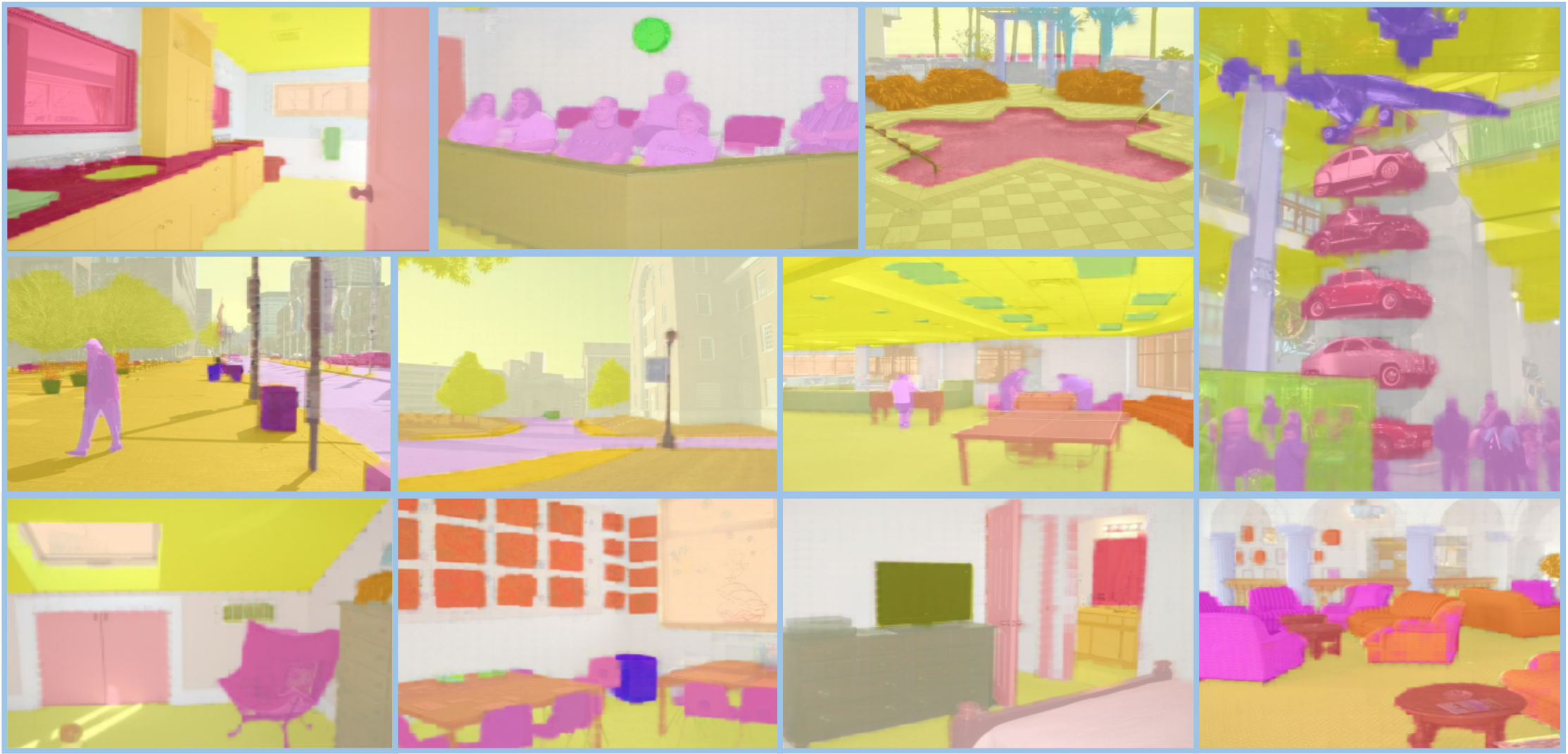}
        \caption{Semantic segmentation on ADE20K}
        \vspace{4pt}
    \end{subfigure}
    \begin{subfigure}{0.85\linewidth}
        \centering
        \setlength{\abovecaptionskip}{0.5pt}
        \includegraphics[width=0.99\linewidth, trim=0 0 0 103, clip]{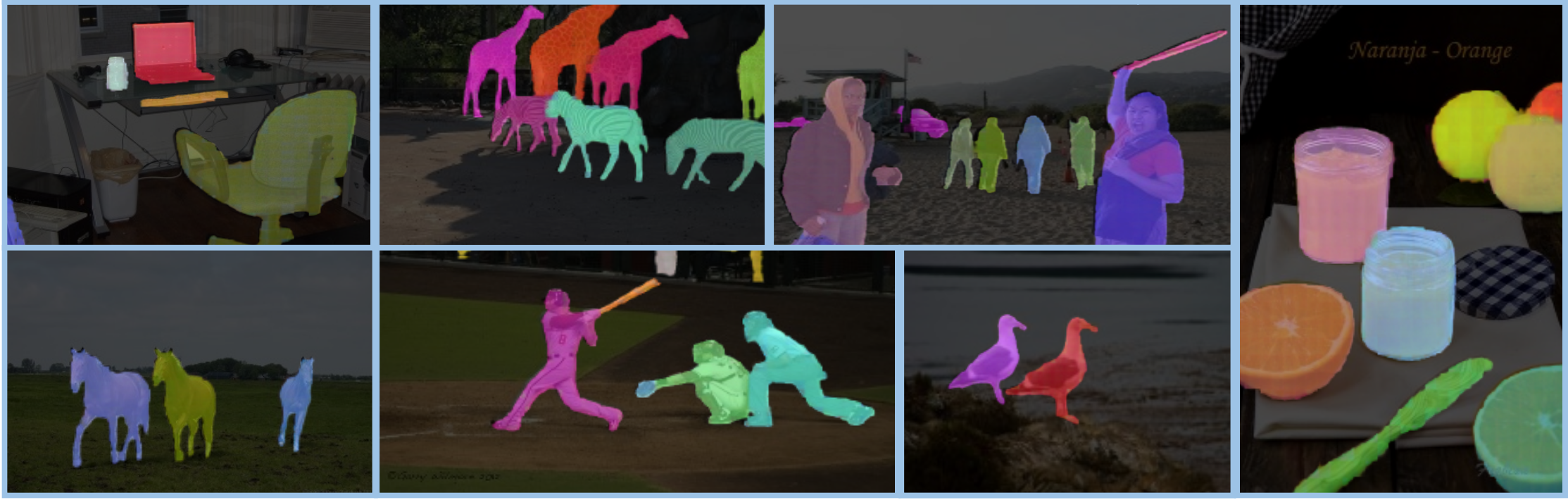}
        \caption{Instance segmentation on COCO}
        \vspace{4pt}
    \end{subfigure}
    \begin{subfigure}{0.85\linewidth}
        \centering
        \setlength{\abovecaptionskip}{0.5pt}
        \includegraphics[width=0.99\linewidth, trim=2 0 3 0, clip]{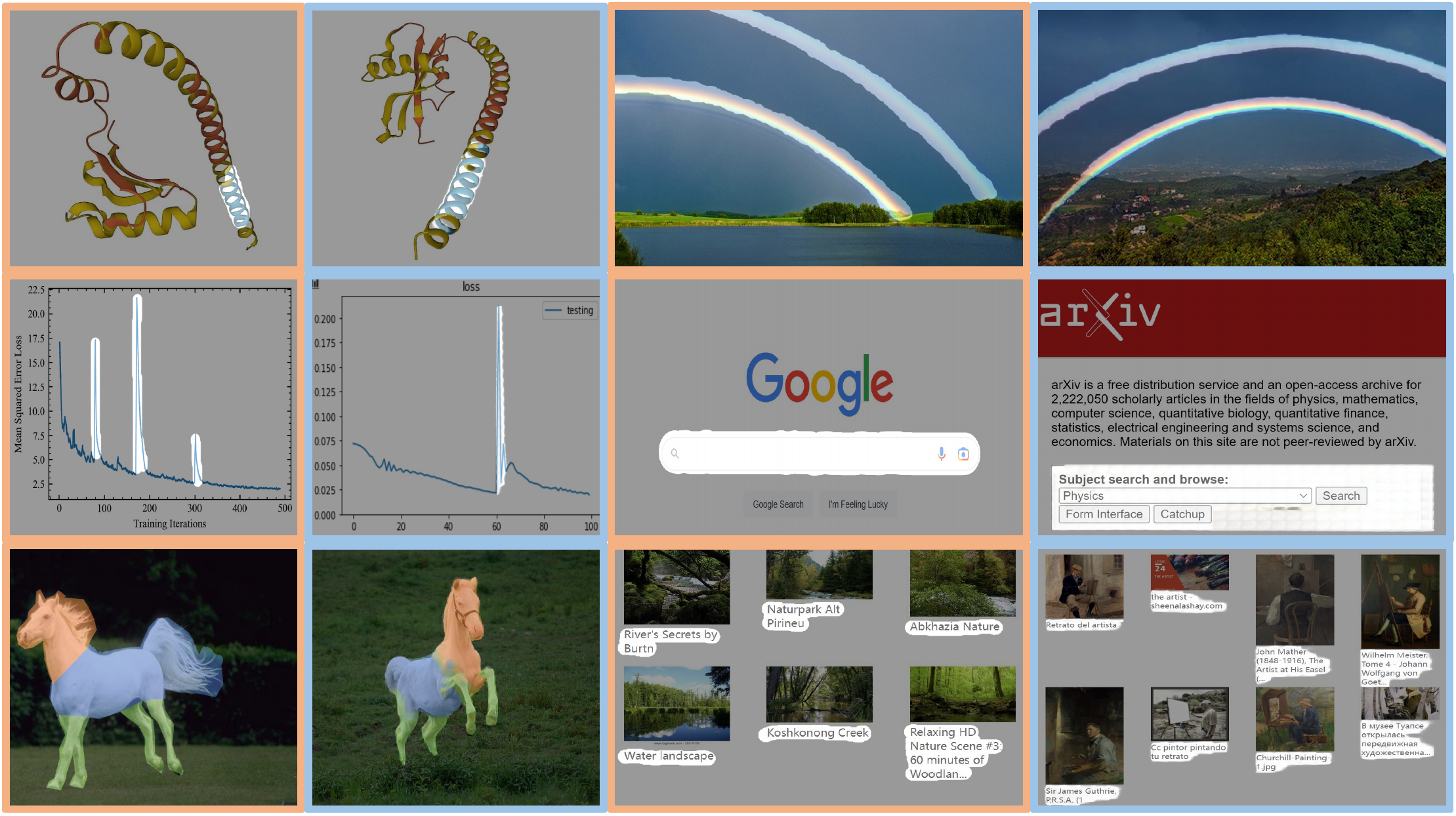}
        \caption{Arbitrary segmentation in the wild}
    \end{subfigure}
    \caption{More examples of \Ours applications. Each test image and the corresponding predicted segmentation are combined for better visualization. For (c), the orange box \textcolor{orange}{$\square$} on the left displays the example/prompt image and its corresponding mask, while the blue box \textcolor{cyan}{$\square$}  on the right shows the input image and the resulting mask output.}
    \label{fig:supp_vis}
\end{figure*}

\myparagraph{Visualizations.}
We provide more visualizations in Figure~\ref{fig:supp_vis}, including semantic segmentation on ADE20K, instance segmentation on COCO, and arbitrary segmentation in the wild.

\begin{table}[th]
\small
\centering
\begin{tabular}{c|cc}
examples
& mIoU    & mAcc    \\
\shline
1
& 18.8    & 27.4    \\
2 
& 25.0    & 34.4    \\
4 
& 28.3    & 37.7    \\
8 
& 30.1    & 38.9    \\
16 
& 31.9    & 40.4    \\
32
& 33.0    & 42.0    \\
tuned
& 39.6    & 50.7   \\
\end{tabular}
\vspace{-1.0em}
\caption{Example-based results on ADE20K semantic segmentation. More examples boost the performance.}
\label{table:supp_ade20k}
\vspace{-1em}
\end{table}

\end{document}